\documentclass{article}

    \usepackage[final]{neurips_2021}

\usepackage[utf8]{inputenc} 
\usepackage[T1]{fontenc}    
\usepackage{hyperref}       
\usepackage{url}            
\usepackage{booktabs}       
\usepackage{amsfonts}       
\usepackage{nicefrac}       
\usepackage{microtype}      
\usepackage{xcolor}         
\usepackage{hyperref}
\hypersetup{
    colorlinks=true,
    urlcolor=blue,
}
\usepackage{dblfloatfix}

\usepackage{amsmath}
\usepackage[pdftex]{graphicx} 
\usepackage{wrapfig}
\setcitestyle{numbers,square,comma}
\usepackage{amsmath}
\DeclareMathOperator*{\argmax}{arg\,max}

\title{Mastering Atari Games with Limited Data}

\author{%
  Weirui Ye\thanks{{\small \texttt{\{ywr20, liush20\}@mails.tsinghua.edu.cn, gaoyangiiis@tsinghua.edu.cn}}}
  \quad
  Shaohuai Liu\footnotemark[1]
  \quad
  Thanard Kurutach\thanks{{\small \texttt{\{thanard.kurutach, pabbeel\}@berkeley.edu}}}
  \quad
  Pieter Abbeel\footnotemark[2]
  \quad
  Yang Gao\footnotemark[1] \,$^\ddagger$ \\
  $^*$Tsinghua University, $^\dagger$UC Berkeley, $^\ddagger$ Shanghai Qi Zhi Institute
}

\begin{document}

\maketitle

\renewcommand{\thefootnote}{\fnsymbol{footnote}}

\begin{abstract}
  Reinforcement learning has achieved great success in many applications. However, sample efficiency remains a key challenge, with prominent methods requiring millions (or even billions) of environment steps to train.  Recently, there has been significant progress in sample efficient image-based RL algorithms; however, consistent human-level performance on the Atari game benchmark remains an elusive goal. We propose a sample efficient model-based visual RL algorithm built on MuZero, which we name EfficientZero. Our method achieves 194.3\% mean human performance and 109.0\% median performance on the Atari 100k benchmark with only two hours of real-time game experience and outperforms the state SAC in some tasks on the DMControl 100k benchmark. This is the first time an algorithm achieves super-human performance on Atari games with such little data. EfficientZero's performance is also close to DQN's performance at 200 million frames while we consume 500 times less data. EfficientZero's low sample complexity \emph{and} high performance can bring RL closer to real-world applicability.
  We implement our algorithm in an easy-to-understand manner and it is available at \url{https://github.com/YeWR/EfficientZero}. We hope it will accelerate the research of MCTS-based RL algorithms in the wider community. 
\end{abstract}

\begin{figure}[!h]
  \begin{center}
  \vskip -.5cm
    \includegraphics[width=\linewidth]{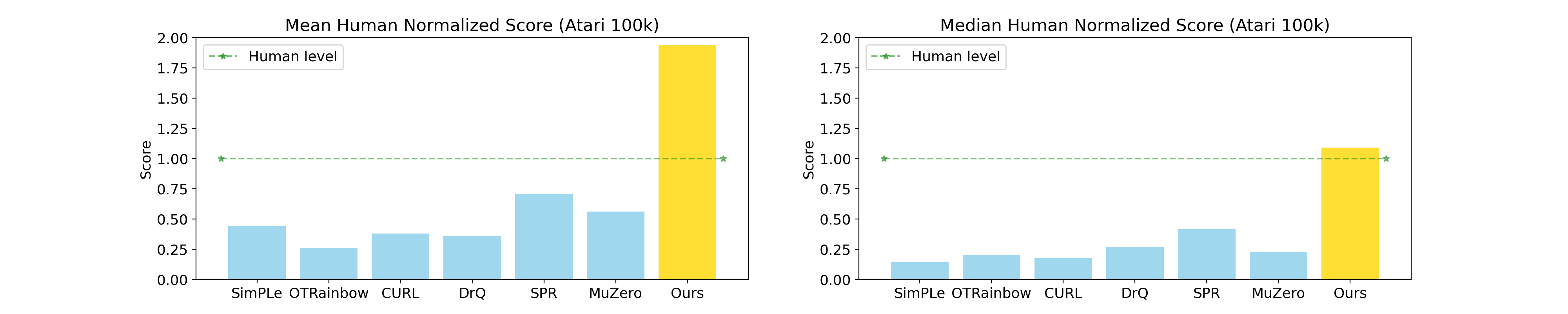}
  \end{center}
  \vskip -0.25cm
  \caption{Our proposed method EfficientZero is 176\% and 163\% better than the previous SoTA performance in mean and median human normalized score and is the first to outperform the average human performance on the Atari 100k benchmark. The high sample efficiency and performance of EfficientZero can bring RL closer to the real-world applications.}
  \label{fig:compare}
  \vskip -0.25cm
\end{figure}

\section{Introduction}\label{sec:intro}
Reinforcement learning has achieved great success on many challenging problems. Notable work includes DQN~\cite{mnih2015human}, AlphaGo~\cite{silver2016mastering} and OpenAI Five~\cite{berner2019dota}. However, most of these works come at the cost of a large number of environmental interactions. For example, AlphaZero ~\cite{silver2017mastering} needs to play 21 million games at training time. On the contrary, a professional human player can only play around 5 games per day, meaning it would take a human player 11,500 years to achieve the same amount of experience. The sample complexity might be less of an issue when applying RL algorithms in simulation and games. However, when it comes to real-world problems, such as robotic manipulation, healthcare, and advertisement recommendation systems, achieving high performance while maintaining low sample complexity is the key to viability. 

People have made a lot of progress in sample efficient RL in the past years~\cite{chua2018deep, deisenroth2011pilco, srinivas2020curl, laskin2020reinforcement, kostrikov2020image, schwarzerdata, kaiser2019model}. Among them, model-based methods have attracted a lot of attention, since both the data from real environments and the ``imagined data'' from the model can be used to train the policy, making these methods particularly sample-efficient~\cite{chua2018deep,deisenroth2011pilco}. However, most of the successes are in state-based environments. In image-based environments, some model-based methods such as MuZero~\cite{schrittwieser2020mastering} and Dreamer V2~\cite{hafner2020mastering} achieve super-human performance, but they are not sample efficient; other methods such as SimPLe~\cite{kaiser2019model} is quite efficient but achieve inferior performance (0.144 human normalized median scores). Recently, data-augmented and self-supervised methods applied to model-free methods have achieved more success in the data-efficient regime~\cite{schwarzerdata}. However, they still fail to achieve the levels which can be expected of a human.

Therefore, for improving the sample efficiency as well as keeping superior performance, we find the following three components are essential to the model-based visual RL agent: a self-supervised environment model,
a mechanism to alleviate the model compounding error, and a method to correct the off-policy issue. 
In this work, we propose EfficientZero, a model-based RL algorithm that achieves high performance with limited data. Our proposed method is built on MuZero. We make three critical changes: (1) use self-supervised learning to learn a temporally consistent environment model, (2) learn the \textit{value prefix} in an end-to-end manner, thus helping to alleviate the compounding error in the model, (3) use the learned model to correct off-policy value targets. 

As illustrated as Figure \ref{fig:compare}, our model achieves state-of-the-art performance on the widely used Atari~\cite{bellemare2013arcade} 100k benchmark and it achieves super-human performance with only 2 hours of real-time gameplay. More specifically, our model achieves 194.3\% mean human normalized performance and 109.0\% median human normalized performance. As a reference, DQN~\cite{mnih2015human} achieves 220\% mean human normalized performance, and 96\% median human normalized performance, at the cost of 500 times more data (200 million frames).
To further verify the effectiveness of EfficientZero, we conduct experiments on some simulated robotics environments of the DeepMind Control (DMControl) suite. It achieves state-of-the-art performance and outperforms the state SAC which directly learns from the ground truth states. Our sample efficient and high-performance algorithm opens the possibility of having more impact on many real-world problems.

\section{Related Work} \label{sec:related}
\subsection{Sample Efficient Reinforcement Learning}
Sample efficiency has attracted significant work in the past. In RL with image inputs, model-based approaches~\cite{hafner2019learning, hafner2019dream} which model the world with both a stochastic and a deterministic component, have achieved promising results for simulated robotic control. \citet{kaiser2019model} propose to use an action-conditioned video prediction model, along with a policy learning algorithm. It achieves the first strong performance on Atari games with as little as 400k frames. However, \citet{kielak2020recent} and \citet{van2019use} argue that this is not necessary to achieve strong results with model-based methods, and they show that when tuned appropriately, Rainbow~\cite{hessel2018rainbow} can achieve comparable results. 

Recent advances in self-supervised learning, such as SimCLR~\cite{chen2020simple}, MoCo~\cite{he2020momentum}, SimSiam~\cite{chen2020exploring} and BYOL~\cite{grill2020bootstrap} have inspired representation learning in image-based RL.
\citet{srinivas2020curl} propose to use contrastive learning in RL algorithms and their work achieves strong performance on image-based continuous and discrete control tasks. Later, \citet{laskin2020reinforcement} and \citet{kostrikov2020image} find that contrastive learning is not necessary, but with data augmentations alone, they can achieve better performance.
Some researchers propose to use contrastive learning to enforce action equivariance on the learned representations \cite{van2020plannable}.
\citet{schwarzerdata} propose a temporal consistency loss, which is combined with data augmentations and achieves state-of-the-art performance. Notably, our self-supervised consistency loss is quite similar to SPR\cite{schwarzerdata}, except we use SimSiam \cite{chen2020simple} while they use BYOL \cite{grill2020bootstrap} as the base self-supervised learning framework. However, 
they only apply the learned representations in a model-free manner, while we combine the learned model with model-based exploration and policy improvement, thus leading to more efficient use of the environment model. 

Despite the recent progress in the sample-efficient RL, today's RL algorithms are still well behind human performance when the amount of data is limited. Although traditional model-based RL is considered more sample efficient than model-free ones, current model-free methods dominate in terms of performance for image-input settings. In this paper, we propose a model-based RL algorithm that for the first time, 
achieves super-human performance on Atari games with limited data. 

\subsection{Reinforcement Learning with MCTS}
Temporal difference learning~\cite{mnih2015human, van2016deep, wang2016dueling, hessel2018rainbow} and policy gradient based methods~\cite{mnih2016asynchronous, lillicrap2015continuous, schulman2015trust, schulman2017proximal} are two types of popular reinforcement learning algorithms. Recently, \citet{silver2016mastering} propose to use MCTS as a policy improvement operator and has achieved great success in many board games, such as Go, Chess, and Shogi~\cite{silver2017mastering}. Later, the algorithm is adapted to learn the world model at the same time~\cite{schrittwieser2020mastering}. It has also been extended to deal with continuous action spaces~\cite{hubert2021learning} and offline data~\cite{schrittwieser2021online}. These MCTS RL algorithms are a hybrid of model-based learning and model-free learning. 

However, most of them are trained with a lot of environmental samples. Our method is built on top of MuZero~\cite{schrittwieser2020mastering}, and we demonstrate that our method can achieve higher sample efficiency while still achieving competitive performance on the Atari 100k benchmark. \citet{de2021visualizing} have studied the potential of using auxiliary loss similar to our self-supervised consistency loss. However, they only test on two low dimensional state-based environments and find the auxiliary loss has mixed effects on the performance. On the contrary, we find that the consistency loss is critical in most environments with high dimensional observations and limited data. 

\subsection{Multi-Step Value Estimation}
In Q-learning~\cite{watkins1989learning}, the target Q value is computed by one step backup. In practice, people find that incorporating multiple steps of rewards at once, i.e. $z_t = \sum_{i=0}^{k-1} \gamma^i  u_{t+i} + \gamma^k v_{t+k}$, where $u_{t+i}$ is the reward from the replay buffer, $v_{t+k}$ is the value estimation from the target network, to compute the value target $z_t$ leads to faster convergence~\cite{mnih2015human,hessel2018rainbow}. However, the use of multi-step value has off-policy issues, since $u_{t+i}$ are not generated by the current policy. In practice, this issue is usually ignored when there is a large amount of data since the data can be thought as approximately on-policy. TD($\lambda$)~\cite{sutton2018reinforcement} and GAE~\cite{schulman2015high} improve the value estimation by better trading off the bias and the variance, but they do not deal with the off-policy issue. 
Recently, image input model-based algorithms such as \citet{kaiser2019model} and
\citet{hafner2019dream} use model imaginary rollouts to avoid the off-policy issue. However, this approach has the risk of model exploitation.
\citet{asadi2019combating} proposed a multi-step model to combat the compounding error.
Our proposed model-based off-policy correction method starts from the rewards in the real-world experience and uses model-based value estimate to bootstrap. Our approach balances between the off-policy issue and model exploitation. 


\section{Background}\label{sec:background}

\subsection{MuZero}

Our method is built on top of the MuZero Reanalyze~\cite{schrittwieser2020mastering} algorithm. For brevity, we refer to it as MuZero throughout the paper. MuZero is a policy learning method based on the Monte-Carlo Tree Search (MCTS) algorithm. The MCTS algorithm operates with an environment model, a prior policy function, and a value function. The environment model is represented as the reward function $\mathcal{R}$ and the dynamic function $\mathcal{G}$: $r_t=\mathcal{R}(s_t, a_t)$, $\hat{s}_{t+1}=\mathcal{G}(s_t, a_t)$, which are needed when MCTS expands a new node. In MuZero, the environment model is learned. Thus the reward and the next state are approximated. Besides, the predicted policy $p_t=$ acts as a search prior over actions of a node. It helps the MCTS focus on more promising actions when expanding the node. MCTS also needs a value function $\mathcal{V}(s_t)$ that measures the expected return of the node $s_t$, which provides a long-term evaluation of the tree's leaf node without further search. MCTS will output an action visit distribution $\pi_t$ over the root node, which is potentially a better policy, compared to the current neural network. Thus, the MCTS algorithm can be thought of as a policy improvement operator. 

In practice, the environment model, policy function, and value function operate on a hidden abstract state $s_t$, both for computational efficiency and ease of environment modeling. The abstract state is extracted by a representation function $\mathcal{H}$ on observations $o_t$: $s_t =\mathcal{H}(o_t)$. All of the mentioned models above are usually represented as neural networks. During training, the algorithm collects roll-out data in the environment using MCTS, resulting in potentially higher quality data than the current neural network policy. The data is stored in a replay buffer. The optimizer minimizes the following loss on the data sampled from the replay buffer:
\begin{equation}
    \begin{aligned}
        \mathcal{L}(u_t, r_t) + \lambda_1 \mathcal{L}(\pi_t, p_t) + \lambda_2 \mathcal{L}(z_t, v_t)
    \end{aligned}
\end{equation}
Here, $u_t$ is the reward from the environment, $r_t=\mathcal{R}(s_t, a_t)$ is the predicted reward, $\pi_t$ is the output visit count distribution of the MCTS, $p_t=\mathcal{P}(s_t)$ is the predicted policy, $z_t = \sum_{i=0}^{k-1} \gamma^i  u_{t+i} + \gamma^k v_{t+k}$ is the bootstrapped value target and $v_t=\mathcal{V}(s_t)$ is the predicted value. Specifically, the reward function $\mathcal{R}$, policy function $\mathcal{P}$, value function $\mathcal{V}$, the representation function $\mathcal{H}$ and the dynamics function $\mathcal{G}$ are trainable neural networks.
It is worth noting that MuZero does not explicitly learn the environment model. Instead, it solely relies on the reward, value, and policy prediction to learn the model. 

\subsection{Monte-Carlo Tree Search}

Monte-Carlo Tree Search~\cite{abramson2014expected,silver2016mastering,silver2017mastering, hafner2020mastering}, or MCTS, is a heuristic search algorithm. In our setup, MCTS is used to find an action policy that is better than the current neural network policy. 

More specifically, MCTS needs an environment model, including the reward function and the dynamics function. It also needs a value function and a policy function, which act as heuristics during search. MCTS operates by expanding a search tree from the current node. It saves computation by selectively expanding nodes. To find a high-quality decision, the expansion process has to balance between exploration versus exploitation, i.e. balance between expanding a node that is promising with more visits versus a node with lower performance but fewer visits. MCTS employs the UCT~\cite{rosin2011multi, kocsis2006bandit} rule, i.e. UCB~\cite{auer2002finite} on trees. At every node expansion step, UCT will select a node as follows~\cite{hafner2020mastering}:
\begin{equation}
    \label{eq:uct}
    \begin{aligned}
        a^k=\argmax_a \left\{ Q(s,a)+P(s,a)\frac{\sqrt{\sum_b N(s, b)}}{1+N(s,a)}\left(c_1 + \log \left(\frac{\sum_b N(s,b)+c_2+1}{c_2}\right)\right) \right\}
    \end{aligned}
\end{equation}
where, $Q(s,a)$ is the current estimate of the Q-value, $P(s,a)$ is the current neural network policy for selecting this action, helping the MCTS prioritize exploring promising part of the tree. During training time, $P(s,a)$ is usually perturbed by noises to allow explorations. $N(s,a)$ denotes how many times this state-action pair is visited in the tree search, and $N(s,b)$ denote that of $a$'s siblings. Thus this term will encourage the search to visit the nodes whose siblings are visited often, but itself less visited. Finally, the last term gives a weights to the previous terms.

After expanding the nodes for a pre-defined number of times, the MCTS will return how many times each action under the root node is visited, as the improved policy to the root node. Thus, MCTS can be considered as a policy improvement operator in the RL setting. 

\section{EfficientZero}
\label{sec:method}
Model-based algorithms have achieved great success in sample-efficient learning from low-dimensional states. However, current visual model-based algorithms either require large amounts of training data or exhibit inferior performance to model-free algorithms in data-limited settings~\cite{schwarzerdata}. Many previous works even suspect whether model-based algorithms can really offer data efficiency when using image observations~\cite{van2019use}. We provide a positive answer here. We propose the EfficientZero, a model-based algorithm built on the MCTS, that achieves super-human performance on the 100k Atari benchmark, outperforming the previous SoTA to a large degree.

When directly running MCTS-based RL algorithms such as MuZero, we find that they do not perform well on the limited-data benchmark. Through our ablations, we confirm the following three issues which pose challenges to algorithms like MuZero in data-limited settings.

\textbf{Lack of supervision on environment model}. First, the learned model in the environment dynamics is only trained through the reward, value and policy functions. 
However, the reward is only a scalar signal and in many scenarios, the reward will be sparse. Value functions are trained with bootstrapping, and thus are noisy. Policy functions are trained with the search process. None of the reward, value and policy losses can provide enough training signals to learn the environment model. 

\textbf{Hardness to deal with aleatoric uncertainty}. Second, we find that even with enough data, the predicted rewards still have large prediction errors. This is caused by the aleatoric uncertainty of the underlying environment. For example, when the environment is hard to model, the reward prediction errors will accumulate when expanding the MCTS tree to a large depth. 

\textbf{Off-policy issues of multi-step value}. Lastly, as for value targets, MuZero uses the multi-step reward observed in the environment. Although it allows rewards to be propagated to the value function faster, it suffers from severe off-policy issues and hinders convergence given limited data. 

To address the above issues, we propose the following three critical modifications, which can greatly improve performance when samples are limited. 

\subsection{Self-Supervised Consistency Loss}

\begin{wrapfigure}{R}{0.4\textwidth}
  \begin{center}
  \vskip -0.25cm
    \includegraphics[width=\linewidth]{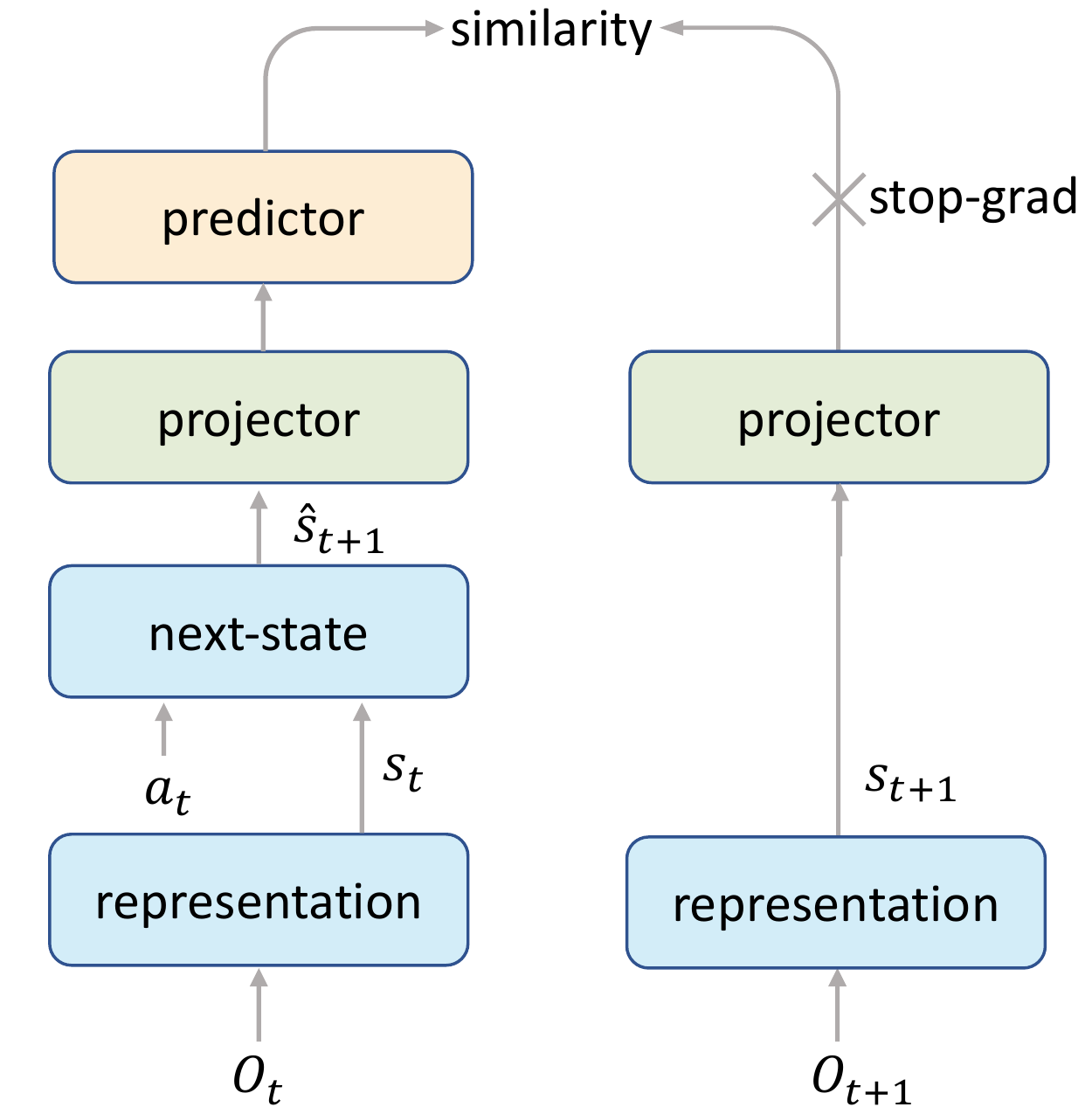}
  \end{center}
  \vskip -0.25cm
  \caption{The self-supervised consistency loss.}
  \label{fig:method-consistency}
  \vskip -0.25cm
\end{wrapfigure}

In previous MCTS RL algorithms, the environment model is either given or only trained with rewards, values, and policies, which cannot provide sufficient training signals due to their scalar nature. The problem is more severe when the reward is sparse or the bootstrapped value is not accurate. The MCTS policy improvement operator heavily relies on the environment model. Thus, it is vital to have an accurate one. 
We notice that the output $\hat{s}_{t+1}$ from the dynamic function $\mathcal{G}$ should be the same as $s_{t+1}$, i.e. the output of the representation function $\mathcal{H}$ with input of the next observation $o_{t+1}$ (Fig. \ref{fig:method-consistency}). This can help to supervise the predicted next state $\hat{s}_{t+1}$ using the actual $s_{t+1}$, which is a tensor with at least a few hundred dimensions. This provides $\hat{s}_{t+1}$ with much more training signals than the default scalar reward and value. 

Notably, in MCTS RL algorithms, the consistency between the hidden states and the predicted states can be shaped through the dynamics function directly without extra models.
More specifically, we adopt the recently proposed SimSiam~\cite{chen2020exploring} self-supervised framework. SimSiam~\cite{chen2020exploring} is a self-supervised method that takes two augmentation views of the same image and pulls the output of the second branch close to that of the first branch, where the first branch is an encoder network without gradient, and the second is the same encoder network with the gradient and a predictor head. The head can simply be an MLP.

Note that SimSiam only learns the representation of individual images, and is not aware of how different images are connected. The learned image representations of SimSiam might not be a good candidate for learning the transition function, since adjacent observations might be encoded to very different representation encodings. We propose a self-supervised method that learns the transition function, along with the image representation function in an end-to-end manner, as Figure \ref{fig:method-consistency} shows. Since we aim to learn the transition between adjacent observations, we pull $o_t$ and $o_{t+1}$ close to each other. The transition function is applied after the representation of $o_t$, such that $s_t$ is transformed to $\hat{s}_{t+1}$, which now represents the same entity as the other branch. Then both of $s_{t+1}$ and $\hat{s}_{t+1}$ go through a common projector network. Since $s_{t+1}$ is potentially a more accurate description of $o_{t+1}$ compared to $\hat{s}_{t+1}$, we make the $o_{t+1}$ branch as the target branch. It is common in self-supervised learning that the second or the third layer from the last is chosen as the features for some reason. Here, we choose the outputs from the representation network or the dynamics network as the hidden states rather than those from the projector or the predictor.
The two adjacent observations provide two views of the same entity. In practice, we find that applying augmentations to observations on the image helps to further improve the learned representation quality~\cite{srinivas2020curl,schwarzerdata}. We also unroll the dynamic function recurrently for 5 further steps and also pull $\hat{s}_{t+k}$ close to $s_{t+k}$ ($k=1,...,5$). Please see 
App.\ref{sec:app-models} 
for more details. We note that our temporal consistency loss is similar to SPR \cite{schwarzerdata}, an unsupervised representation method applied on rainbow. However, the consistency loss in our case is applied in a model-based manner, and we use the SimSiam loss function.

\subsection{End-To-End Prediction of the Value Prefix}
In model-based learning, the agent needs to predict the future states conditioned on the current state and a series of hypothetical actions. The longer the prediction, the harder to predict it accurately, due to the compounding error in the recurrent rollouts. This is called the state aliasing problem. The environment model plays an important role in MCTS. The state aliasing problem harms the MCTS expansion, which will result in sub-optimal exploration as well as sub-optimal action search. 

\begin{wrapfigure}{R}{0.5\textwidth}
  \begin{center}
  \vskip -.5cm
    \includegraphics[width=\linewidth]{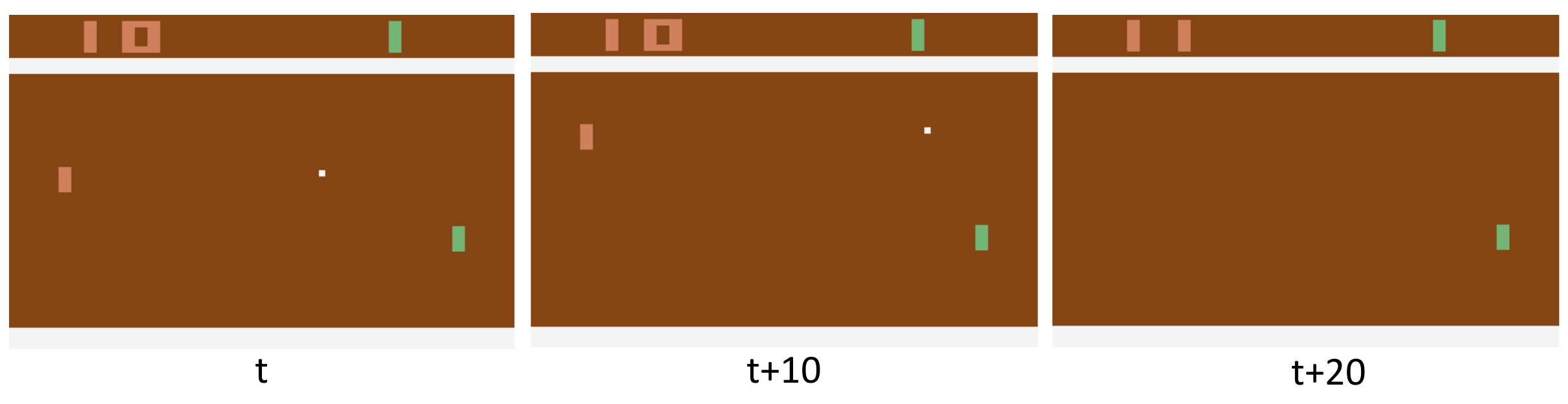}
  \end{center}
  \vskip -0.25cm
  \caption{A sample trajectory from the Atari Pong game. In this case, the right player didn't move and missed the ball. }
  \label{fig:pong}
  \vskip -0.25cm
\end{wrapfigure}

Predicting the reward from an aliased state is a hard problem. For example, as shown in Figure \ref{fig:pong}, the right agent loses the ball. If we only see the first observation, along with future actions, it is very hard both for an agent and a human to predict at which exact future timestep the player would lose a point. However, it is easy to predict the agent will miss the ball after a sufficient number of timesteps if he does not move. In practice, a human will never try to predict the exact step that he loses the point but will imagine over a longer horizon and thus get a more confident prediction. 

Inspired by this intuition, we propose an end-to-end method to predict the \textit{value prefix}. We notice that the predicted reward is always used in the estimation of the Q-value $Q(s, a)$ in UCT of Equation \ref{eq:uct}
\begin{equation}
\label{eq:qsa}
Q(s_t, a)=\sum_{i=0}^{k-1} \gamma^i r_{t+i} + \gamma^k v_{t+k}
\end{equation}
, where $r_{t+i}$ is the reward predicted from unrolled state $\hat{s}_{t+i}$. We name the sum of rewards $\sum_{i=0}^{k-1} \gamma^i r_{t+i}$ as the value prefix, since it is used as a prefix in the later Q-value computation. 

We propose to predict value prefix from the unrolled states ($s_t, \hat{s}_{t+1}, \cdots, \hat{s}_{t+k-1}$) in an end-to-end manner, i.e. $\text{value-prefix}=f(s_t, \hat{s}_{t+1}, \cdots, \hat{s}_{t+k-1})$. Here $f$ is some neural network architecture that takes in a variable number of inputs and outputs a scalar. We choose the LSTM in our experiment. During the training time, the LSTM is supervised at every time step, since the value prefix can be computed whenever a new state comes in. This per-step rich supervision allows the LSTM can be trained well even with limited data. Compared with the naive per step reward prediction and summation approach, the end-to-end value prefix prediction is more accurate, because it can automatically handle the intermediate state aliasing problem. See Experiment Section \ref{sec:value_prefix_proof} for empirical evaluations. As a result, it helps the MCTS to explore better, and thus increases the performance. See 
App.\ref{sec:app-models} 
for architectural details. 

\subsection{Model-Based Off-Policy Correction}
\label{sec:off_policy}
In MCTS RL algorithms, the value function fits the value of the current neural network policy. However, in practice as MuZero Reanalyze does, the value target is computed by sampling a trajectory from the replay buffer and computing:
$z_t = \sum_{i=0}^{k-1} \gamma^i  u_{t+i} + \gamma^k v_{t+k}$. This value target suffers from off-policy issues, since the trajectory is rolled out using an older policy, and thus the value target is no longer accurate. When data is limited, we have to reuse the data sampled from a much older policy, thus exaggerating the inaccurate value target issue.

In previous model-free settings, there is no straightforward approach to fix this issue. On the contrary, since we have a model of the environment, we can use the model to imagine an "online experience". More specifically, we propose to use rewards of a dynamic horizon $l$ from the old trajectory, where $l<k$ and $l$ should be smaller if the trajectory is older. This reduces the policy divergence by fewer rollout steps. Further, we redo an MCTS search with the current policy on the last state $s_{t+l}$ and compute the empirical mean value at the root node. This effectively corrects the off policy issue using imagined rollouts with current policy and reduces the increased bias caused by setting $l$ less than $k$. Formally, we propose to use the following value target:
\begin{equation}
\label{eq:off_policy}
    z_t = \sum_{i=0}^{l-1} \gamma^i  u_{t+i}  + \gamma^l \nu^{\text{MCTS}}_{t+l}
\end{equation}
where $l<=k$  and the older the sampled trajectory, the smaller the $l$. $\nu^{\text{MCTS}}(\text{s}_{t+l})$ is the root value of the MCTS tree expanded from $s_{t+l}$ with the current policy, as MuZero non-Reanalyze does.
See 
App.\ref{sec:app-train_details} 
for how to choose $l$.
In practice, the computation cost of the correction is two times on the reanalyzed side. However, the training will not be affected due to the parallel implementation. 
\section{Experiments}
In this section, we aim to evaluate the sample efficiency of the proposed algorithm. Here, the sample efficiency is measured by the performance of each algorithm at a common, small amount of environment transitions, i.e. the better the performance, the higher the sample efficiency. More specifically, we use the Atari 100k benchmark. Intuitively, this benchmark asks the agent to learn to play Atari games within two hours of real-world game time. Additionally, we conduct some ablation studies to investigate and analyze each component on Atari 100k. To further show the sample efficiency, we apply EfficientZero to some simulated robotics environments on the DMControl 100k benchmark, which contains the same 100k environment steps.

\subsection{Environments}
\textbf{Atari 100k} Atari 100k was first proposed by the SimPLe \cite{kaiser2019model} method, and is now used by many sample-efficient RL works, such as \citet{srinivas2020curl,laskin2020reinforcement,kostrikov2020image, schwarzerdata}. The benchmark contains 26 Atari games, and the diverse set of games can effectively measure the performance of different algorithms. The benchmark allows the agent to interact with 100 thousand environment steps, i.e. 400 thousand frames due to a frameskip of 4, with each environment. 100k steps roughly correspond to 2 hours of real-time gameplay, which is far less than the usual RL settings. For example, DQN \cite{mnih2015human} uses 200 million frames, which is around 925 hours of real-time gameplay. Note that the human player's performance is tested after allowing the human to get familiar with the game after 2 hours as well. 
We report the raw performance on each game, as well as the mean and median of the human normalized score. The human normalized score is defined as: $(\text{score}_\text{agent}-\text{score}_\text{random})/(\text{score}_\text{human}-\text{score}_\text{random})$. 

We compare our method to the following baselines. 
(1) SimPLe \cite{kaiser2019model}, a model-based RL algorithm that learns an action conditional video prediction model and trains PPO within the learned environment. (2) OTRainbow \cite{kielak2020recent}, which tunes the hyper-parameters of the Rainbow \cite{hessel2018rainbow} method to achieve higher sample efficiency. (3) CURL \cite{srinivas2020curl}, which uses contrastive learning as a side task to improve the image representation quality. (4) DrQ \cite{kostrikov2020image}, which adds data augmentations to the input images while learning the original RL objective. (5) SPR \cite{schwarzerdata}, the previous SoTA in Atari 100k which proposes to augment the Rainbow \cite{hessel2018rainbow} agent with data augmentations as well as a multi-step consistency loss using BYOL-style self-supervision. (6) MuZero \cite{schrittwieser2020mastering} 
with our implementations and the same hyper-parameters as EfficientZero.
(7) Random Agent (8) Human performance. 

\textbf{DeepMind Control 100k} \citet{tassa2018deepmind} propose the DMControl suite, which includes some challenging visual robotics tasks with continuous action space. 
And some works \cite{hafner2019dream, srinivas2020curl} have benchmarked for the sample efficiency on the DMControl 100k which contains 100k environment steps data. Since the MCTS-based methods cannot deal with tasks with continuous action space, we discretize each dimension into 5 discrete slots in MuZero \cite{schrittwieser2020mastering} and EfficientZero. To avoid the dimension explosion, we evaluate EfficientZero in three low-dimensional tasks. 

We compare our method to the following baselines. 
(1) Pixel SAC, which applies SAC directly to pixels. (2) SAC-AE \cite{yarats2019improving}, which combines the SAC and an auto-encoder to handle image-based inputs. (3) State SAC, which applies SAC directly to ground truth low dimensional states rather than the pixels. (4) Dreamer \cite{hafner2019dream}, which learns a world model and is trained in dreamed scenarios. (5) CURL \cite{srinivas2020curl}, the previous SoTA in DMControl 100k. (6) MuZero \cite{schrittwieser2020mastering} with action discretizations. 

\subsection{Results}
Table \ref{tab:atari_results_full} shows the results of EfficientZero on the Atari 100k benchmark. Normalizing our score with the score of human players, EfficientZero achieves a mean score of 1.904 and a median score of 1.160. As a reference, DQN \cite{mnih2015human} achieves a mean and median performance of 2.20 and 0.959 on these 26 games. However, it is trained with 500 times more data (200 million frames). 
For the first time, an agent trained with only 2 hours of game data can outperform the human player in terms of the mean and median performance. Among all games, our method outperforms the human in 14 out of 26 games. Compared with the previous state-of-the-art method (SPR \cite{schwarzerdata}), we are 170\% and 180\% better in terms of mean and median score respectively.
As for more robust results, we record the aggregate metrics in 
App.\ref{app:evaluation}
with statistical tools proposed by \citet{agarwal2021deep}.

Apart from the Atari games, EffcientZero achieves remarkable results in the simulated tasks. As shown in Table \ref{tab:dmc_full}, EffcientZero outperforms CURL, the previous SoTA, to a considerable degree and keeps a smaller variance but MuZero cannot work well. Notably, EfficientZero achieves comparable results to the state SAC, which consumes the ground truth states and is considered as the oracles. 

\begin{table}[ht]
    \caption{Scores on the Atari 100k benchmark (3 runs with 32 seeds). EfficientZero achieves super-human performance with only 2 hours of real-time game play. Our method is 176\% and 163\% better than the previous SoTA performance, in mean and median human normalized score respectively.}
    \label{tab:atari_results_full}
\begin{center}
\begin{small}
\centering
\scalebox{0.85}{
\centering
\begin{tabular}{lccccccccr}
\toprule
Game &                  Random &    Human &   SimPLe &OTRainbow&     CURL &      DrQ &     SPR & MuZero & Ours\\
\midrule
Alien               &    227.8 &   7127.7 &    616.9 &   \textbf{824.7} &    558.2 &    771.2 &   801.5 & 530.0 & 808.5 \\
Amidar              &      5.8 &   1719.5 &     88.0 &    82.8 &    142.1 &    102.8 &   \textbf{176.3} & 38.8 & 148.6 \\
Assault             &    222.4 &    742.0 &    527.2 &   351.9 &    600.6 &    452.4 &   571.0 & 500.1 & \textbf{1263.1} \\
Asterix             &    210.0 &   8503.3 &   1128.3 &   628.5 &    734.5 &    603.5 &   977.8 & 1734.0 & \textbf{25557.8} \\
Bank Heist          &     14.2 &    753.1 &     34.2 &   182.1 &    131.6 &    168.9 &   \textbf{380.9} & 192.5 & 351.0 \\
BattleZone          &   2360.0 &  37187.5 &   5184.4 &  4060.6 &  14870.0 &  12954.0 & \textbf{16651.0} & 7687.5 & 13871.2 \\
Boxing              &      0.1 &     12.1 &      9.1 &     2.5 &      1.2 &      6.0 &    35.8 & 15.1 & \textbf{52.7} \\
Breakout            &      1.7 &     30.5 &     16.4 &     9.8 &      4.9 &     16.1 &    17.1 & 48.0 & \textbf{414.1} \\
ChopperCmd           &    811.0 &   7387.8 &   1246.9 &  1033.3 &   1058.5 &    780.3 &   974.8 & \textbf{1350.0} & 1117.3 \\
Crazy Climber       &  10780.5 &  35829.4 &  62583.6 & 21327.8 &  12146.5 &  20516.5 & 42923.6 & 56937.0 & \textbf{83940.2} \\
Demon Attack        &    152.1 &   1971.0 &    208.1 &   711.8 &    817.6 &    1113.4&   545.2 & 3527.0 & \textbf{13003.9} \\
Freeway             &      0.0 &     29.6 &     20.3 &    25.0 &     \textbf{26.7} &      9.8 &    24.4 & 21.8 & 21.8 \\
Frostbite           &     65.2 &   4334.7 &    254.7 &   231.6 &   1181.3 &    331.1 &  \textbf{1821.5} & 255.0 & 296.3 \\
Gopher              &    257.6 &   2412.5 &    771.0 &   778.0 &    669.3 &    636.3 &   715.2 & 1256.0 & \textbf{3260.3} \\
Hero                &   1027.0 &  30826.4 &   2656.6 &  6458.8 &   6279.3 &   3736.3 &  7019.2 & 3095.0 & \textbf{9315.9} \\
Jamesbond           &     29.0 &    302.8 &    125.3 &   112.3 &    471.0 &    236.0 &   365.4 & 87.5 & \textbf{517.0} \\
Kangaroo            &     52.0 &   3035.0 &    323.1 &   605.4 &    872.5 &    940.6 &  \textbf{3276.4} & 62.5 & 724.1 \\
Krull               &   1598.0 &   2665.5 &   4539.9 &  3277.9 &   4229.6 &   4018.1 &  3688.9 & 4890.8 & \textbf{5663.3} \\
Kung Fu Master      &    258.5 &  22736.3 &  17257.2 &  5722.2 &  14307.8 &   9111.0 & 13192.7 & 18813.0 & \textbf{30944.8} \\
Ms Pacman           &    307.3 &   6951.6 &   \textbf{1480.0} &   941.9 &   1465.5 &    960.5 &  1313.2 & 1265.6 & 1281.2 \\
Pong                &    -20.7 &     14.6 &     12.8 &     1.3 &    -16.5 &     -8.5 &    -5.9 & -6.7 & \textbf{20.1} \\
Private Eye         &     24.9 &  69571.3 &     58.3 &   100.0 &    \textbf{218.4} &    -13.6 &   124.0 & 56.3 & 96.7 \\
Qbert               &    163.9 &  13455.0 &   1288.8 &   509.3 &   1042.4 &    854.4 &   669.1 & 3952.0 & \textbf{13781.9} \\
Road Runner         &     11.5 &   7845.0 &   5640.6 &  2696.7 &   5661.0 &   8895.1 & 14220.5 & 2500.0 & \textbf{17751.3} \\
Seaquest            &     68.4 &  42054.7 &    683.3 &   286.9 &    384.5 &    301.2 &   583.1 & 208.0 & \textbf{1100.2} \\
Up N Down           &    533.4 &  11693.2 &   3350.3 &  2847.6 &   2955.2 &   3180.8 & \textbf{28138.5} & 2896.9 & 17264.2 \\
\midrule
Normed Mean         &    0.000 &    1.000 &    0.443 &    0.264 &   0.381 &    0.357 &   0.704 & 0.562 & \textbf{1.943} \\
Normed Median       &    0.000 &    1.000 &    0.144 &    0.204 &   0.175 &    0.268 &   0.415 & 0.227 & \textbf{1.090} \\
\bottomrule
\end{tabular}
}
\end{small}
\end{center}
\end{table}

\begin{table}[ht]
    \caption{Scores achieved by EfficientZero (mean \& standard deviation for 10 seeds) and some baselines on some low-dimensional environments on the DMControl 100k benchmark. EfficientZero achieves state-of-art performance and comparable results to the state-based SAC. 
    }
    \label{tab:dmc_full}
\begin{center}
\begin{small}
\centering
\scalebox{0.85}{
\centering
\begin{tabular}{lccccccccr}
\toprule
Task & CURL & Dreamer & MuZero & SAC-AE & Pixel SAC & State SAC & EfficientZero \\
\midrule
Cartpole, Swingup & 582$\pm$ 146 & 326$\pm$27 & 218.5 $\pm$ 122 & 311$\pm$11 & 419$\pm$40 & 835$\pm$22 & \textbf{813$\pm$19} \\
Reacher, Easy & 538$\pm$ 233 & 314$\pm$155 & 493 $\pm$ 145 & 274$\pm$14 & 145$\pm$30 & 746$\pm$25 & \textbf{952$\pm$34} \\
Ball in cup, Catch & 769$\pm$ 43 & 246 $\pm$ 174 & 542 $\pm$ 270 & 391$\pm$ 82 & 312$\pm$ 63 & 746$\pm$91 & \textbf{942$\pm$17} \\
\bottomrule
\end{tabular}
}
\end{small}
\end{center}
\end{table}

\subsection{Ablations}\label{sec:ablations}
In Section \ref{sec:method}, we discuss three issues that prevent MuZero from achieving high performance when data is limited: (1) the lack of environment model supervision, (2) the state aliasing issue, and (3) the off-policy target value issue. We propose three corresponding approaches to fix those issues and demonstrate the usefulness of the combination of those approaches on a wide range of 26 Atari games. In this section, we will analyze each component individually.

\begin{table}[ht]
    \caption{Ablations of the self-supervised consistency, end-to-end value prefix and model-based off-policy correction. We remove one component at a time and evaluate the corresponding version on the 26 Atari games. Each component matters and the consistency one is the most significant. The detailed results are attached in 
    App.\ref{app:more_ablation}
    .}
    \label{tab:atari_ablations_full_overall}
\begin{center}
\begin{small}
\centering
\scalebox{0.85}{
\centering
\begin{tabular}{lcccc}
\toprule
Game &                  Full &    w.o. consistency &   w.o. value prefix & w.o. off-policy correction\\
\midrule
Normed Mean & \textbf{1.943} & 0.881 & 1.482 & 1.475 \\
Normed Median & \textbf{1.090} & 0.340 & 0.552 & 0.836 \\
\bottomrule
\end{tabular}
}
\end{small}
\end{center}
\end{table}

\textbf{Each Component} Firstly, we do an ablation study by removing the three components from our full model one at a time. 
As shown in Table \ref{tab:atari_ablations_full_overall}, we find that removing any one of the three components will lead to a performance drop compared to our full model. Furthermore, the richer learning signals are the aspect Muzero lacks most in the low-data regime as the largest performance drop is from the version without consistency supervision. As for the performance in the high-data regime, We find that the temporal consistency can significantly accelerate the training. The value prefix seems to be helpful during the early learning process, but not as much in the later stage. The off-policy correction is not necessary as it is specifically designed under limited data.

\begin{figure}[t]
  \begin{center}
  \vskip -.5cm
    \includegraphics[width=\linewidth]{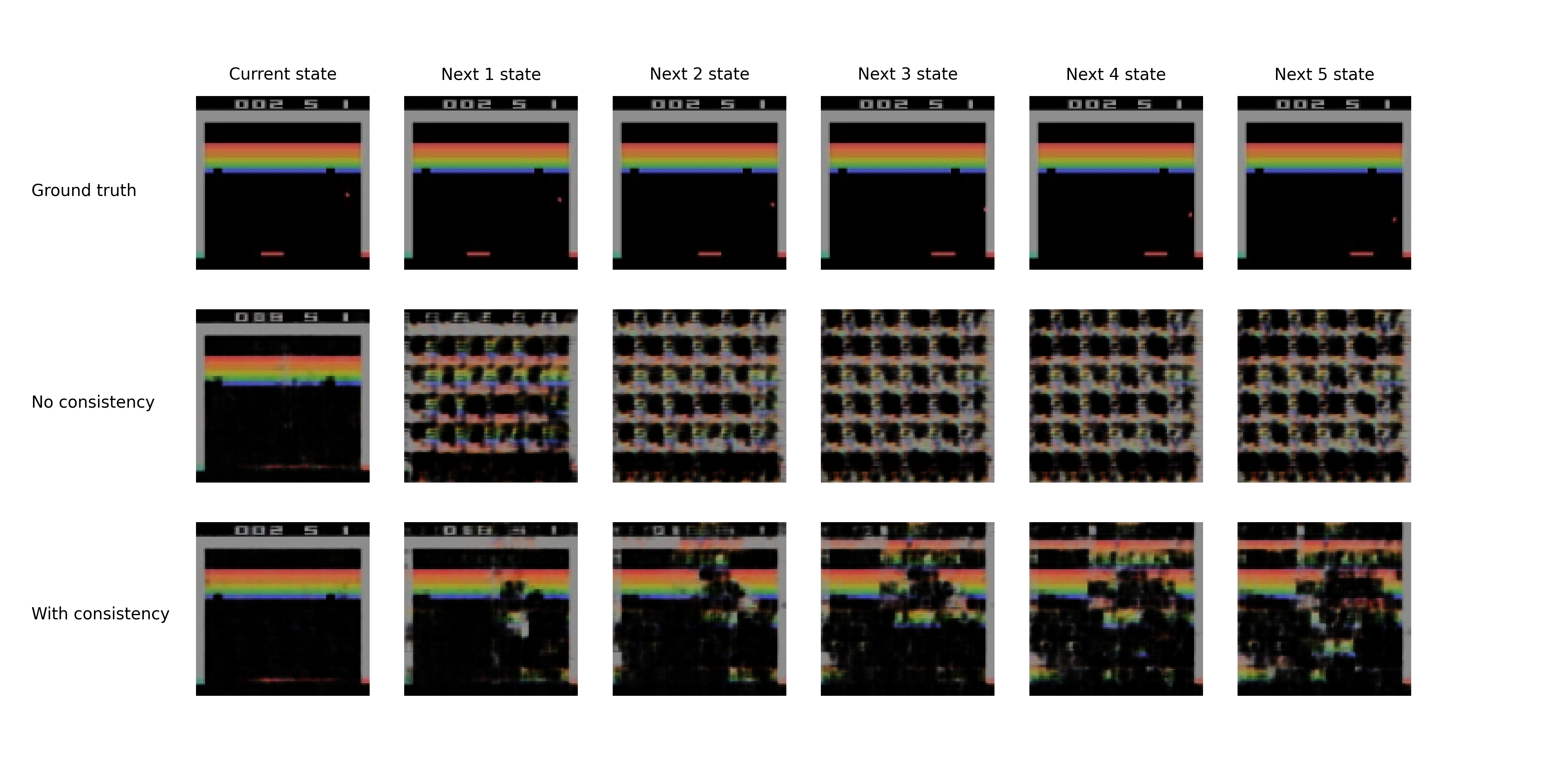}
  \end{center}
  \vskip -0.25cm
  \caption{Evaluations of image reconstructions based on latent states extracted from the model with or without self-supervised consistency. The predicted next states with consistency can basically be reconstructed into observations while the ones without consistency cannot.}
  \label{fig:ablation-consistency}
  \vskip -0.25cm
\end{figure}

\textbf{Temporal Consistency} As the version without self-supervised consistency cannot work well in most of the games, we attempt to dig into the reason for such phenomenon. We design a decoder $\mathcal{D}$ to reconstruct the original observations, taking the latent states as inputs. Specifically, the architecture of $\mathcal{D}$ and the $\mathcal{H}$ are symmetrical, which means that all the convolutional layers are replaced by deconvolutional layers in $\mathcal{D}$ and the order of the layers are reversed in $\mathcal{D}$. Therefore, $\mathcal{H}$ is an encoder to obtain state $s_t$ from observation $o_t$ and $\mathcal{D}$ tries to decode the $o_t$ from $s_t$. In this ablation, we freeze all parameters of the trained EfficientZero network with or without consistency respectively and the reconstructed results are shown in different columns of Figure \ref{fig:ablation-consistency}. We regard the decoder as a tool to visualize the current states and unrolled states, shown in different rows of Figure \ref{fig:ablation-consistency}. Here we note that $\mathcal{M}_{\text{con}}$ is the trained EfficientZero model with consistency and $\mathcal{M}_{\text{non}}$ is the one without consistency.
As shown in Figure \ref{fig:ablation-consistency}, as for the current state $s_t$, the observation is reconstructed well enough in the two versions. However, it is remarkable that the the decoder given $\mathcal{M}_{\text{non}}$ can not reconstruct images from the unrolled predicted states $\hat{s}_{t+k}$ while the one given  $\mathcal{M}_{\text{con}}$ can reconstruct basic observations.

\begin{wrapfigure}{R}{0.65\textwidth}
  \begin{center}
  \vskip -.5cm
    \includegraphics[width=\linewidth]{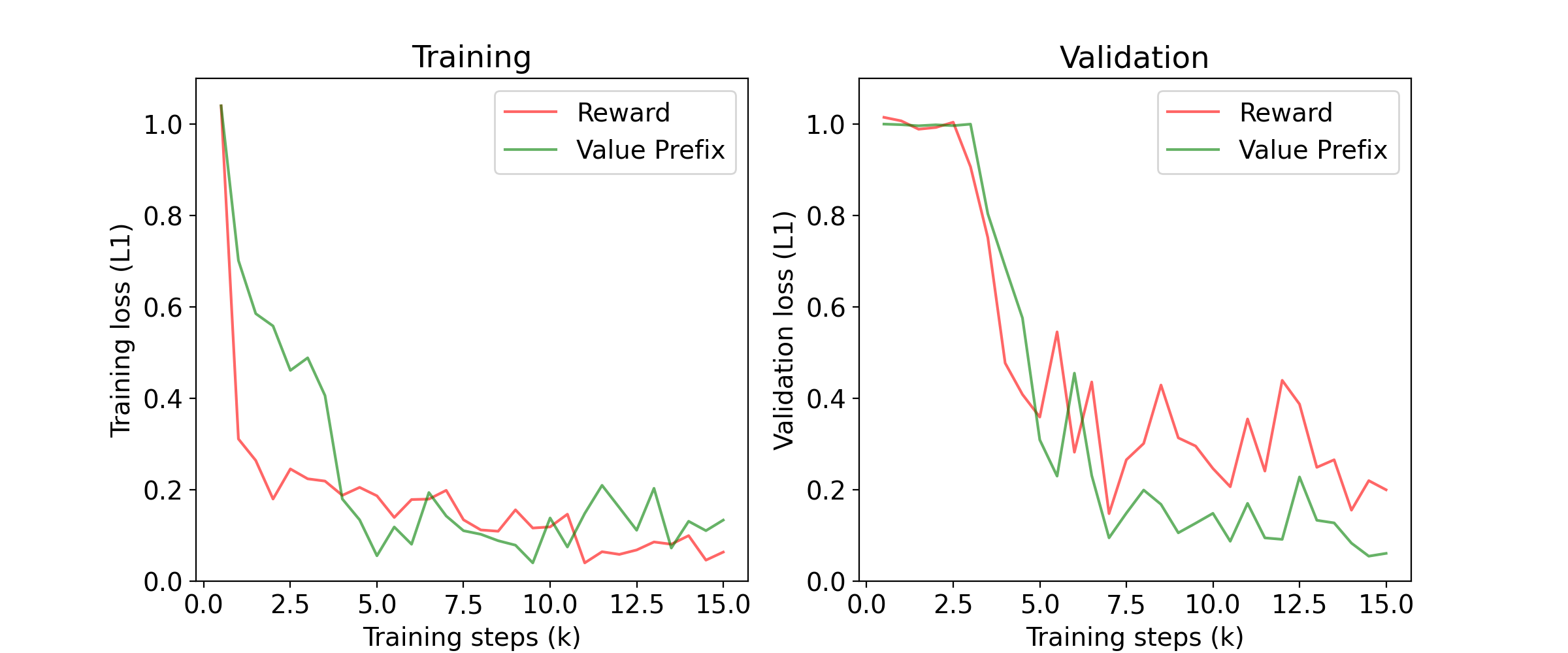}
  \end{center}
  \vskip -0.25cm
  \caption{Training and validation losses of direct reward prediction method and the value prefix method.}
  \label{fig:value-prefix}
  \vskip -0.25cm
\end{wrapfigure}

To sum up, there are some distributional shifts between the latent states from the representation network and the states from the dynamics function without consistency. The consistency component can reduce the shift and provide more supervision for training the dynamics network.

\textbf{Value Prefix} \label{sec:value_prefix_proof}
We further validate our assumptions in the end-to-end learning of value prefix, i.e. the state aliasing problem will cause difficulty in predicting the reward, and end-to-end learning of value prefix can alleviate this phenomenon. 

To fairly compare directly predicting the reward versus end-to-end learning of the value prefix, we need to control for the dataset that both methods are trained on.
Since during the RL training, the dataset distribution is determined by the method, we opt to load a half-trained Pong model and rollout total 100k steps as the common static dataset. 
We split this dataset into a training set and a validation set. Then we run both the direct reward prediction and the value prefix method on the training split. 

As shown in Figure \ref{fig:value-prefix}, we find that the direct reward prediction method has lower losses on the training set. However, the value prefix's validation error is much smaller when unrolled for 5 steps. This shows that the value prefix method avoids overfitting the hard reward prediction problem, and thus it can reduce the state aliasing problem, reaching a better generalization performance.

\textbf{Off-Policy Correction} To prove the effectiveness of the off-policy correction component, we compare the error between the target values and the ground truth values with or without off-policy correction. Specifically, the ground truth values are estimated by Monte Carlo sampling.

We train a model for the game UpNDown with total 100k training steps, and collect the trajectories at different training stages respectively (20k, 40k, ..., 100k steps). Then we calculate the ground truth values with the final model. We choose the trajectories at the same stage (20k) and use the final model to evaluate the target values with or without off-policy correction, following the Equation \ref{eq:off_policy}.
We evaluate the L1 error of the target values and the ground truth, as shown in Table \ref{tab:correction}. The error of unrolled next 5 states means the average error of the unrolled 1-5 states with dynamics network from current states.
The error is smaller in both current states and the unrolled states with off-policy correction. Thus, the correction component does reduce the bias caused by the off-policy issue.

\begin{table}[h]
    \caption{Ablations of the off-policy correction: L1 error of the target values versus the ground truth values. Take UpNDown as an example.}
    \label{tab:correction}
\begin{center}
\begin{small}
\centering
\scalebox{1.0}{
\centering
\begin{tabular}{lccccccccr}
\toprule
States & Current state & Unrolled next 5 states (Avg.) & All states (Avg.)  \\
\midrule
Value error without correction & 0.765 & 0.636 & 0.657 \\
Value error with correction & \textbf{0.533} & \textbf{0.576} & \textbf{0.569} \\
\bottomrule
\end{tabular}
}
\end{small}
\end{center}
\end{table}

Furthermore, we also ablate the value error of the trajectories at distinct stages in Table \ref{tab:correction_2}.
We can find that the value error becomes smaller as the trajectories are fresher. This indicates that the off-policy issue is severe due to the staleness of the data. More significantly, the off-policy correction can provide more accurate target value estimation for the trajectories at distinct time-steps as all the errors with correction shown in the table are smaller than those without correction at the same stage. 

\begin{table}[h]
    \caption{Ablations of the off-policy correction: Average L1 error of the values of the trajectories at distinct stages. Take UpNDown as an example.}
    \label{tab:correction_2}
\begin{center}
\begin{small}
\centering
\scalebox{1.0}{
\centering
\begin{tabular}{lccccccccr}
\toprule
Stages of trajectories & 20k & 40k & 60k & 80k & 100k  \\
\midrule
Value error without correction & 0.657 & 0.697 & 0.628 & 0.574 & 0.441 \\
Value error with correction & \textbf{0.569} & \textbf{0.552} & \textbf{0.537} & \textbf{0.488} & \textbf{0.397}  \\
\bottomrule
\end{tabular}
}
\end{small}
\end{center}
\end{table}

\section{Discussion}
\label{sec:discussion}
In this paper, we propose a sample-efficient model-based method EfficientZero. It achieves super-human performance on the Atari games with as little as 2 hours of the gameplay experience and state-of-the-art performance on some DMControl tasks. Apart from the full results, we do detailed ablation studies to examine the effectiveness of the proposed components.
This work is one step towards running RL in the physical world with complex sensory inputs. In the future, we plan to extend it to more directions, such as a better design for the continuous action space. And we also plan to study the acceleration of MCTS and how to combine this framework with life-long learning.

\begin{ack}
This work is supported by the Ministry of Science and Technology of the People's Republic of China, the 2030 Innovation Megaprojects ``Program on New Generation Artificial Intelligence'' (Grant No. 2021AAA0150000).
\end{ack}

\bibliographystyle{IEEEtran}
\bibliography{main}
\appendix

\section{Appendix}
\subsection{Models and Hyper-parameters}
\label{sec:app-models}
As for the architecture of the networks, there are three parts in our model pipeline: the representation part, the dynamics part, and the prediction part. The architecture of the representation part is as follows:
\begin{itemize}
    \item 1 convolution with stride 2 and 32 output planes, output resolution 48x48. (BN + ReLU)
    \item 1 residual block with 32 planes.
    \item 1 residual downsample block with stride 2 and 64 output planes, output resolution 24x24.
    \item 1 residual block with 64 planes.
    \item Average pooling with stride 2, output resolution 12x12. (BN + ReLU)
    \item 1 residual block with 64 planes.
    \item Average pooling with stride 2, output resolution 6x6. (BN + ReLU)
    \item 1 residual block with 64 planes.
\end{itemize}
, where the kernel size is $3 \times 3$ for all operations.

As for the dynamics network, we follow the architecture of MuZero \cite{schrittwieser2020mastering} but reduce the residual blocks from 16 to 1. Furthermore, we add an extra residual link in the dynamics part to keep the information of historical hidden states during recurrent inference. The design of the dynamics network is listed here:
\begin{itemize}
    \item Concatenate the input states and input actions into 65 planes.
    \item 1 convolution with stride 2 and 64 output planes. (BN)
    \item A residual link: add up the output and the input states. (ReLU)
    \item 1 residual block with 64 planes.
\end{itemize}

In the prediction part, we use two-layer MLPs with batch normalization to predict the reward, value, or policy. Considering the stability of the prediction part, we set the weights and bias of the last layer to zero in prediction networks. As for the reward prediction network, it predicts the sum of the rewards, namely value prefix: $r_{t}, h_{t+1} = \mathcal{R}(\hat{s}_{t+1}, h_t)$, where $r_t$ is the predicted sum of rewards, $h_0$ is zero-initialized and hidden size of LSTM is 512. The architecture of the value prediction network is as follows:
\begin{itemize}
    \item 1 1x1convolution and 16 output planes. (BN + ReLU)
    \item Flatten.
    \item LSTM with 512 hidden size. (BN + ReLU)
    \item 1 fully connected layers and 32 output dimensions. (BN + ReLU)
    \item 1 fully connected layers and 601 output dimensions.
\end{itemize}
The horizontal length of the LSTM during training is limited to the unrolled steps $l_{\text{unroll}}=5$, but it will be larger in MCTS as the dynamics process can go deeper. Therefore, we reset the hidden state of LSTM after $\zeta = 5$ steps of recurrent inference, where $\zeta$ is the valid horizontal length. 

The design of the reward and policy prediction networks are the same except for the dimension of the outputs:
\begin{itemize}
    \item 1 residual block with 64 planes.
    \item 1 1x1convolution and 16 output planes. (BN + ReLU)
    \item Flatten.
    \item 1 fully connected layers and 32 output dimensions. (BN + ReLU)
    \item 1 fully connected layers and $D$ output dimensions.
\end{itemize}
, where $D=601$ in the reward prediction network and $D$ is equal to the action space in the policy prediction network.

\begin{table}[t]
    \caption{Hyper-parameters for EfficientZero on Atari games}
    \label{tab:param}
\begin{center}
\begin{small}
\centering
\scalebox{1.0}{
\centering
\begin{tabular}{lc}
\toprule
Parameter & Setting \\
\midrule
Observation down-sampling & 96 $\times$ 96 \\
Frames stacked & 4 \\
Frames skip & 4 \\
Reward clipping & True \\
Terminal on loss of life & True \\
Max frames per episode & 108K \\
Discount factor & $0.997^4$ \\
Minibatch size & 256 \\
Optimizer & SGD \\
Optimizer: learning rate & 0.2 \\
Optimizer: momentum & 0.9 \\
Optimizer: weight decay ($c$) & 0.0001 \\
Learning rate schedule & 0.2 $\rightarrow$ 0.02 \\
Max gradient norm & 5 \\
Priority exponent ($\alpha$) & 0.6 \\
Priority correction ($\beta$) & 0.4 $\rightarrow$ 1 \\
Training steps & 120K \\
Evaluation episodes & 32 \\
Min replay size for sampling & 2000 \\
Self-play network updating inerval & 100 \\ 
Target network updating interval & 200 \\
Unroll steps ($l_{\text{unroll}}$) & 5 \\
TD steps ($k$) & 5 \\
Policy loss coefficient ($\lambda_1$) & 1 \\
Value loss coefficient ($\lambda_2$) & 0.25 \\
Self-supervised consistency loss coefficient ($\lambda_3$) & 2 \\
LSTM horizontal length ($\zeta$) & 5 \\
Dirichlet noise ratio ($\xi$) & 0.3 \\
Number of simulations in MCTS ($N_{\text{sim}}$) & 50 \\
Reanalyzed policy ratio & 0.99 \\
\bottomrule
\end{tabular}
}
\end{small}
\end{center}
\end{table}

Here is the brief introduction of the training pipeline, taking one-step rollout as an example.
\begin{equation}
    \begin{aligned}
        s_t &= \mathcal{H}(o_t) \\
        s_{t+1} &= \mathcal{H}(o_{t+1}) \\
        \hat{s}_{t+1} &= \mathcal{G}(s_t, a_t) \\
        v_{t} &= \mathcal{V}(s_t) \\
        p_{t} &= \mathcal{P}(s_t) \\
        r_{t}, h_{t+1} &= \mathcal{R}(\hat{s}_{t+1}, h_t)= \mathcal{R}(\mathcal{G}(s_t, a_t), h_t) \\
    \end{aligned}
\end{equation}
, where $\mathcal{H}$ is the representation network, $\mathcal{G}$ is the dynamics network, $\mathcal{V}$ is the value prediction network, $\mathcal{P}$ is the policy prediction network, $\mathcal{R}$ is the reward (\textit{value prefix}) prediction network. $o_t, s_t, a_t$ are observations, states and actions. $h_t$ is the hidden states in recurrent neural networks.

Here is the training loss, taking one-step rollout as an example:
\begin{equation}
    \begin{aligned}
        \mathcal{L}_{\text{similarity}}(s_{t+1}, \hat{s}_{t+1}) &= \mathcal{L}_2(sg({P}_1(s_{t+1})), P_2(P_1(\hat{s}_{t+1}))) \\
        \mathcal{L}_t(\theta) &= \mathcal{L}(u_t, r_t) + \lambda_1 \mathcal{L}(\pi_t, p_t) + \lambda_2 \mathcal{L}(z_t, v_t) \\ 
        & + \lambda_3 \mathcal{L}_{\text{similarity}}(s_{t+1}, \hat{s}_{t+1}) + c ||\theta||^2 \\
        \mathcal{L}(\theta) &= \frac{1}{l_{\text{unroll}}} \sum_{i=0}^{l_{\text{unroll}}-1} \mathcal{L}_{t+i}(\theta)
    \end{aligned}
\end{equation}
, where $\mathcal{L}$ is the total loss of the unrolled $l_{\text{unroll}}$ steps, $\mathcal{L}_{1}$ is the Cross-Entropy loss, and $\mathcal{L}_2$ is the negtive cosine similarity loss. Besides, $P_1$ is a 3-layer MLP while $P_2$ is a 2-layer MLP. The dimension of the hidden layers is 512 and the dimension of the output layers is 1024. We add batch normalization between every two layers in those MLP except the final layer. $sg({P}_1)$ means stopping gradients.

We stack 4 historical frames, with an interval of 4 frames-skip. Thus the input effectively covers 16 frames of the game history. We stack the input images on the channel dimension, resulting in a $96\times 96 \times 12$ tensor. We do not use any extra state normalization besides the batch norm and we choose reward clipping to keep better scales in the searching process.

Generally, compared with MuZero \cite{schrittwieser2020mastering}, we reduce the number of residual blocks and the number of planes as we find that there is no capability issue caused by much smaller networks in our EfficientZero with limited data. In another word, such a tiny network can acquire good performance in the limited setting. 

For other details, we provide hyper-parameters in Table \ref{tab:param}. It is notable that we train the model for 120k steps where we only collect data during the first 100k steps. 
In this way, the latter trajectories can be fully used in training. Besides, the learning rate will drop after every 100k training steps (from 0.2 to 0.02 at 100k).

\subsection{More Ablations}
\label{app:more_ablation}
In the experiment section 
, we list some ablation studies to prove the effectiveness of each component. In this section, we will display more results for the ablation study.

Firstly, the detailed results of the ablation study of each component are listed in Table \ref{tab:atari_ablations_full}. In this table, We find that the full version of EfficientZero outperforms the others without any one of the components. Furthermore, for those environments EfficientZero can already solve, the performance is similar between the full version and the version without off-policy correction, such as Breakout, Pong, etc. In such a case, the off-policy issue is not severe, which is the reason for this phenomenon.
Besides, for some environments with sparse rewards, the value prefix component matters, such as Pong; and for those with dense rewards, the state aliasing problem has less negative effects for the reward signals are sufficient, such as Qbert. As for the version without self-supervised consistency, the results of all the environments are much poorer.

\begin{table}[ht]
    \caption{Ablations of the self-supervised consistency, end-to-end value prefix and model-based off-policy correction on more Atari games. (Scores on the Atari 100k benchmark)}
    \label{tab:atari_ablations_full}
\begin{center}
\begin{small}
\centering
\scalebox{0.85}{
\centering
\begin{tabular}{lcccc}
\toprule
Game &                  Full &    w.o. consistency &   w.o. value prefix & w.o. off-policy correction\\
\midrule
Alien & 808.5 & \textbf{961.3} & 558 & 619.4 \\
Amidar & 148.6 & 32.2 & 31.0 & \textbf{256.3} \\
Assault & \textbf{1263.1} & 572.9 & 955.0 & 1190.4 \\
Asterix & \textbf{25557.8} & 2065.6 & 7330.0 & 13525.0 \\
Bank Heist & \textbf{351.0} & 165.6 & 273.0 & 297.5 \\
BattleZone & 13871.2 & 14063.0 & 9900.0 & \textbf{16125.0} \\
Boxing & 52.7 & 6.1 & \textbf{60.2} & 30.5 \\
Breakout & \textbf{414.1} & 237.4 & 379.2 & 400.3 \\
ChopperCommand & 1117.3 & 1138.0 & 1280 & \textbf{1487.5} \\
Crazy Climber & 83940.2 & 75550.0 & \textbf{106090.0} & 70681.0 \\
Demon Attack & \textbf{13003.9} & 5973.8 & 6818.5 & 8640.6 \\
Freeway & \textbf{21.8} & 21.8 & 21.8 & 21.8 \\
Frostbite & \textbf{296.3} & 248.8 & 235.2 & 227.5 \\
Gopher & \textbf{3260.3} & 1155 & 2792.0 & 2275.0 \\
Hero & \textbf{9315.9} & 5824.4 & 3167.5 & 9053.0 \\
Jamesbond & \textbf{517.0} & 154.7 & 380.0 & 356.3 \\
Kangaroo & \textbf{724.1} & 375.0 & 200.0 & 687.5 \\
Krull & \textbf{5663.3} & 4178.625 & 4527.6 & 3635.6 \\
Kung Fu Master & \textbf{30944.8} & 19312.5 & 25980.0 & 25025.0 \\
Ms Pacman & 1281.2 & 1090.0 & \textbf{1475.0} & 1297.2 \\
Pong & \textbf{20.1} & -1.5 & 16.8 & 19.5 \\
Private Eye & 96.7 & \textbf{100.0} & 100.0 & 100.0 \\
Qbert & \textbf{13781.9} & 5340.7 & 6360.0 & 13637.5 \\
Road Runner & \textbf{17751.3} & 2700.0 & 3010.0 & 9856.0 \\
Seaquest & \textbf{1100.2} & 460.0 & 468.0 & 843.8 \\
Up N Down & \textbf{17264.2} & 3040.0 & 7656.0 & 4897.2 \\
\midrule
Normed Mean & \textbf{1.943} & 0.881 & 1.482 & 1.475 \\
Normed Median & \textbf{1.090} & 0.340 & 0.552 & 0.836 \\
\bottomrule
\end{tabular}
}
\end{small}
\end{center}
\end{table}

In addition, we do the ablation study for the data augmentation technique in the consistency component to examine the effect of data augmentations.  
We apply a random small shift of 0-4 pixels as well as the change of the intensity as the augmentation techniques.
Here we choose several Atari games and train the model for 100k steps. The results are shown in Table \ref{tab:ablation-augmentation}. We can find that the version without data augmentation has similar performances while the version without consistency component is worse. This indicates that the improvement of the consistency component is basically from the self-supervised learning loss rather than the data augmentation.

\begin{table}[ht]
    \caption{Ablations of the data augmentation technique in the consistency component. Results show that the data augmentation has limited improvement in EfficientZero and the self-supervised training loss is more significant.}
    \label{tab:ablation-augmentation}
\begin{center}
\begin{small}
\centering
\scalebox{0.85}{
\centering
\begin{tabular}{lccc}
\toprule
Game &                  Full &    w.o. consistency &   w.o. data augmentation \\
\midrule
Asterix & 6218.8 & 1350.0 & \textbf{13884.0} \\
Breakout & \textbf{388.8} & 12.0 & 365.2 \\
Demon Attack & \textbf{10536.6} & 5973.8 & 8730.0 \\
Gopher & \textbf{2828.8} & 1155.0 & 1823.75 \\
Pong & \textbf{19.8} & -8.5 & 13.9 \\
Qbert & \textbf{15268.8} & 2304.7 & 14286.0 \\
Seaquest & \textbf{1321.0} & 460.0 & 1125.0 \\
Up N Down & 10238.1 & 3040.0 & \textbf{16380.0} \\
\bottomrule
\end{tabular}
}
\end{small}
\end{center}
\end{table}

Finally, we also do the ablation study for the MCTS root value and the dynamic horizon in the off-policy correction component. Here we choose several Atari games and train the model for 100k steps. As shown in Table \ref{tab:ablation-off_policy}, the version without dynamic horizon has poorer results than that without the MCTS root value. In the off-policy correction component, the dynamic horizon seems more important.

\begin{table}[ht]
    \caption{Ablations of the techniques (the MCTS root value and the dynamic horizon) in the off-policy correction component. The dynamic horizon seems more important than the MCTS root value when data is limited.}
    \label{tab:ablation-off_policy}
\begin{center}
\begin{small}
\centering
\scalebox{0.85}{
\centering
\begin{tabular}{lcccc}
\toprule
Game &                  Full &    w.o. off-policy correction &   w.o. dynamic horizon & w.o. MCTS root value \\
\midrule
Asterix & 6218.8 & 2706.3 & 3263.0 & \textbf{6288.0} \\
Breakout & 388.8 & \textbf{468.6} & 427.0 & 387.8 \\
Demon Attack & \textbf{10536.6} & 8640.6 & 9211.1 & 10063.0 \\
Gopher & \textbf{2828.8} & 2275.0 & 2459.2 & 2651.0 \\
Pong & \textbf{19.8} & 19.5 & 19.2 & 14.5 \\
Qbert & \textbf{15268.8} & 3948.4 & 7945 & 14738.0 \\
Seaquest & \textbf{1321.0} & 1248.0 & 1292.0 & 876.0 \\
Up N Down & \textbf{10238.1} & 3240.0 & 4772.0 & 9925.6 \\
\bottomrule
\end{tabular}
}
\end{small}
\end{center}
\end{table}

\subsection{MCTS Details}

Our policy searching approach is based on Monte-Carlo tree search (MCTS). We follow the procedure in MuZero \cite{schrittwieser2020mastering}, which includes three stages and repeats the searching process for $N_{\text{sim}}=50$ simulations. Here are some brief introductions for each stage.

\textbf{Selection} In the selection part, it targets at choosing an appropriate unvisited node while balancing exploration and exploitation with UCT:
\begin{equation}
    \label{eq:uct_full}
    \begin{aligned}
        a^k= \left\{ \argmax_a Q(s,a)+P(s,a)\frac{\sqrt{\sum_b N(s, b)}}{1+N(s,a)}\left(c_1 + \log \left(\frac{\sum_b N(s,b)+c_2+1}{c_2}\right)\right) \right\}
    \end{aligned}
\end{equation}
, where $Q(s, a)$ is the average Q values after simulations, $N(s, a)$ is the total visit counts at state $s$ by selecting action $a$, and $P(s, a)$ is the policy prior set in the expansion process. In each simulation, the MCTS starts from the root node $s^0$. And for each time-step $k = 1...l$ of the simulation, the algorithm will select the action $a^k$ according to the UCT.  Usually, $c_1=1.25$ and $c_2=19652$ according to the literature~\cite{silver2016mastering,silver2017mastering,hafner2020mastering}. 

However, the default Q value of the unvisted node is set to 0, which indicates the worst state. To give a better Q-value estimation of the unvisited nodes, we evaluate a mean Q value mechanism in each simulation for tree nodes, similar to the implementation of Elf OpenGo \cite{tian2019elf}.
\begin{equation}
    \begin{aligned}
        \hat{Q}(s^{\text{root}}) &= 0 \\
        \hat{Q}(s) &= \frac{\hat{Q}(s^{\text{parent}}) + \sum_{b}\mathbf{1}_{N(s,b) > 0}Q(s, b)}{1 + \sum_{b}\mathbf{1}_{N(s,b) > 0}} \\
        Q(s, a) :&=
        \left\{
            \begin{array}{rcl}
            Q(s, a)     &      & {N(s, a) > 0}\\
            \hat{Q}(s)       &      & {N(s, a) = 0}
            \end{array} 
        \right. 
    \end{aligned}
\end{equation}
, where $\hat{Q}(s)$ is the estimated Q value for unvisited nodes to make better selections considering exploration and exploitation. $s^{\text{root}}$ is the state of the root node and $s^{\text{parent}}$ is the state of the parent node of $s$. In experiments, we find that the mean Q value mechanism gives a better exploration than the default one.

\textbf{Expansion} Then the newly selected node will be expanded with the predicted reward and policy as its prior. Furthermore, when the root node is to expand, we apply the Dirichlet noise to the policy prior during the self-play stage and the reanalyzing stage to give more explorations.
\begin{equation}
    P(s, a) := (1 - \rho) P(s, a) + \rho \mathcal{N}_{\mathcal{D}}(\xi)
\end{equation}
, where $\mathcal{N}_{\mathcal{D}}(\xi)$ is the Dirichlet noise distribution, $\rho, \xi$ is set to 0.25 and 0.3 respectively. However, we do not use any noise and set $\rho$ to 0 instead for those non-root node or during evaluations.

\textbf{Backup}
After selecting and expanding a new node, we need to backup along the current searching trajectory to update the $Q(s, a)$. Considering the scales of values in distinct environments, we compute a normalized Q-value by using the minimum-maximum values calculated along with all visited tree nodes, which is applied in MuZero\cite{schrittwieser2020mastering}. However, when the data is limited, the small difference between the minimum and maximum values will result in overconfidence in UCT calculation. For example, when all the Q-values in those visited tree nodes are in a range of 0 to $10^{-4}$, the normalized Q-value of $10^{-5}$ and $5 \times 10^{-5}$ will make a huge difference as one is normalized to $0.1$ and another is $0.5$. Therefore, we set a threshold here to reduce overconfidence in such occasions, which is called the soft minimum-maximum updates:
\begin{equation}
    \begin{aligned}
        \Bar{Q}(s^{k-1}, a^k) = \frac{Q(s^{k-1}, a^k) - \min_{(s, a) \in Tree}Q(s, a)}{\max(\max_{(s, a) \in Tree}Q(s, a) - \min_{(s, a) \in Tree}Q(s, a), \epsilon)}
    \end{aligned}
\end{equation}
, where $\epsilon$, the threshold to give a smooth range of the min-max bound, is set to $0.01$.

After all the expansions in the MCTS, we will obtain average value and visit count distributions of the root node. Here, the root value can be applied in off-policy correction and the visit count distribution is the target policy distribution:
\begin{equation}
    \pi(s, a) = \frac{N(s, a)^{1/T}}{\sum_{b}N(s, b)^{1/T}}
\end{equation}

We decay the temperature of the MCTS output policy distribution here twice during training, at 50\% and 75\% of the training progress to 0.5 and 0.25 respectively.

\subsection{Training Details}
\label{sec:app-train_details}
In this subsection, we will introduce more training details.

\begin{figure}
  \begin{center}
  \vskip -.5cm
    \includegraphics[width=1\linewidth]{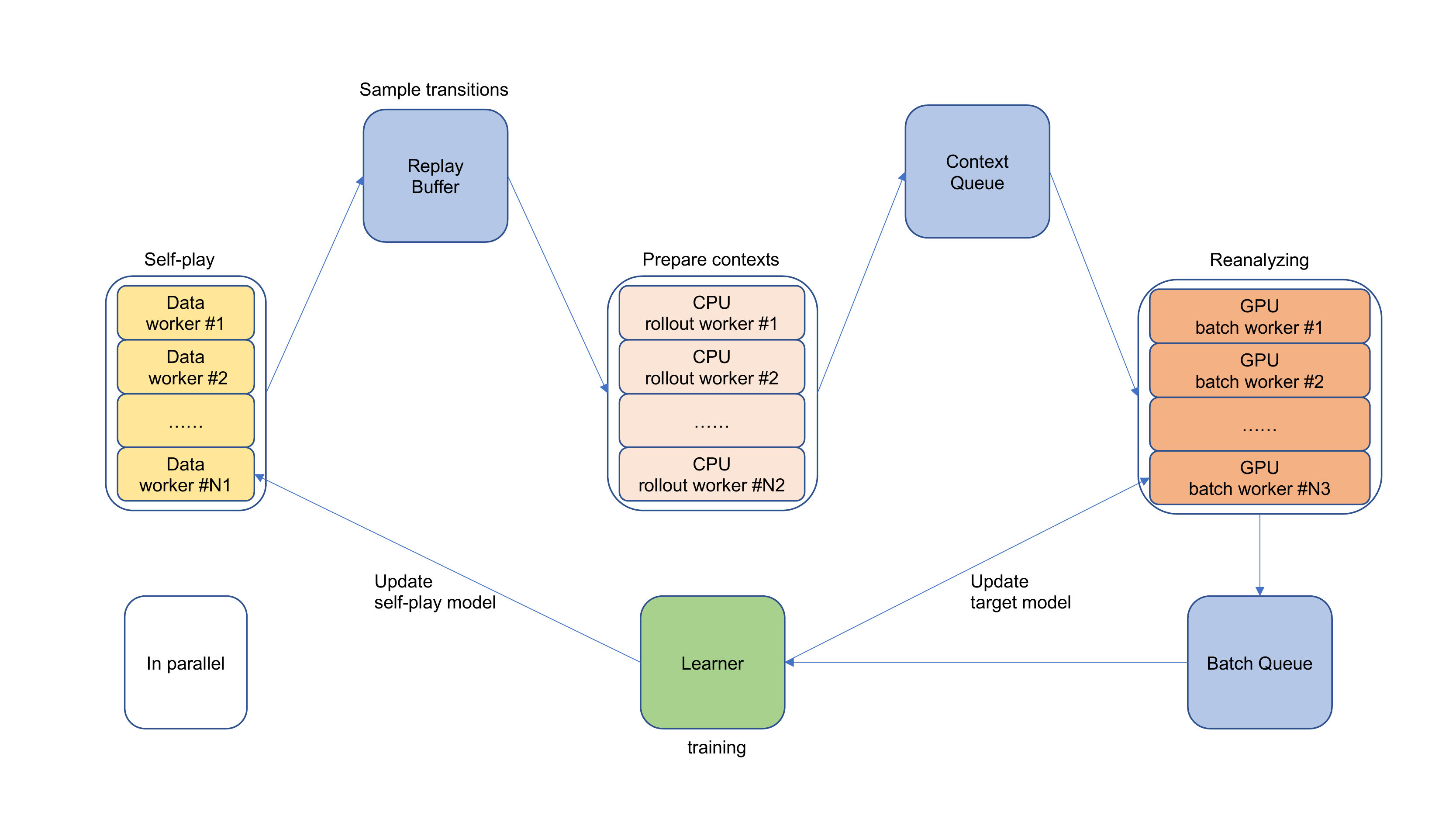}
  \end{center}
  \vskip -0.25cm
  \caption{Pipeline of the EfficientZero implementation.}
  \label{fig:pipeline}
  \vskip -0.25cm
\end{figure}

\textbf{Pipeline} As for the code implementation of EfficientZero, we design a paralleled architecture with a double buffering mechanism in Pytorch and Ray, as shown in Figure \ref{fig:pipeline}. 

Intuitively, we will describe the training process in a synchronized way. Firstly, the data workers called self-play actors are aimed at doing self-play with the given model updated within 600 training steps and then they will send the rolled-out trajectories into the replay buffer. Then the CPU rollout workers attempt to prepare the contexts of those batch transitions sampled from the replay buffer, in which way only CPU resources are required. Afterward, the GPU batch workers reanalyze those past data with the given contexts by the given target model, and most of the time-consuming parts in this procedure are in GPUs. Considering the frequent utilization of CPUs and GPUs in MCTS, the searching process is assigned for those GPU workers. Finally, the learner will obtain the reanalyzed batch and begin to train the agent.

The learner, all the data workers, CPU workers, and GPU workers start in parallel. The data workers and CPU workers share the replay buffer to sample data while the CPU and GPU workers share a context queue for reanalyzing data. Besides, the learner and the GPU workers use a batch queue to communicate. In such a design, we can utilize the CPU and GPU as much as possible.

\textbf{Self-play} 
During self-play, the priorities of the transition to collect are set to the max of the whole priorities in replay buffer.
We also update the priority in EfficientZero according to MuZero \cite{schrittwieser2020mastering}:
$P(i) = \frac{p_i^{\alpha}}{\sum_{k}p_k^{\alpha}}$, where $p_i$ is the L1 error of the value during training. And the we scale with important sampling ratio 
$w_i = (\frac{1}{N \times P(i)})^{\beta}$. We set $\alpha$ to 0.6 and anneal $\beta$ from $0.4$ to $1.0$, following prioritized replay \cite{schaul2015prioritized}. However, we find the priority mechanism only improves a little with limited data.
Considering the long horizons in atari games, we collect the intermediate sequences of 400 moves. 

\textbf{Reanalyze} The reanalyzed part is introduced in MuZero \cite{schrittwieser2020mastering}, which revisits the past trajectories and re-executes the data with lasted target model to obtain a fresher value and policy with model inference as well as MCTS.

For the off-policy correction, the target values are reanalyzed as follows:
\begin{equation}
    \begin{aligned}
        z_t &= \sum_{i=0}^{l-1} \gamma^i  u_{t+i}  + \gamma^l \nu^{\text{MCTS}}_{t+l}, \\
        l &= (k - \lfloor \frac{T_{\text{current}} - T_{s_t}}{\tau T_{\text{total}}} \rfloor).\text{clip}(1, k), l \in [1, k]
    \end{aligned}
\end{equation}
, where $k$ is the TD steps here, and is set to 5; $T_{\text{current}}$ is the current training steps, $T_{s_t}$ is the training steps of collecting the data $s_t$, $T_{\text{total}}$ is the total training steps (100k), and $\tau$ is a coefficient which is set to 0.3. Intuitively, $l$ is to define how fresh the collected data $s_t$ is. When the trajectory is stale, we need to unroll less to estimate the target values for the sake of the gaps between current model predictions and the stale trajectory rollouts. Besides, we replace the predicted value $v_{t+k}$ with the averaged root value from MCTS $\nu^{\text{MCTS}}_{t+l}$ to alleviate the off-policy bias.

Notably, we re-sample Dirichlet noise into the MCTS procedure in reanalyzed part to improve the sample efficiency with a more diverse searching process. Besides, we reanalyze the policy among 99\% of the data and reanalyze the value among 100\% data.

\subsection{Evaluation}
\label{app:evaluation}
\begin{figure}
  \begin{center}
  \vskip -.5cm
    \includegraphics[width=1\linewidth]{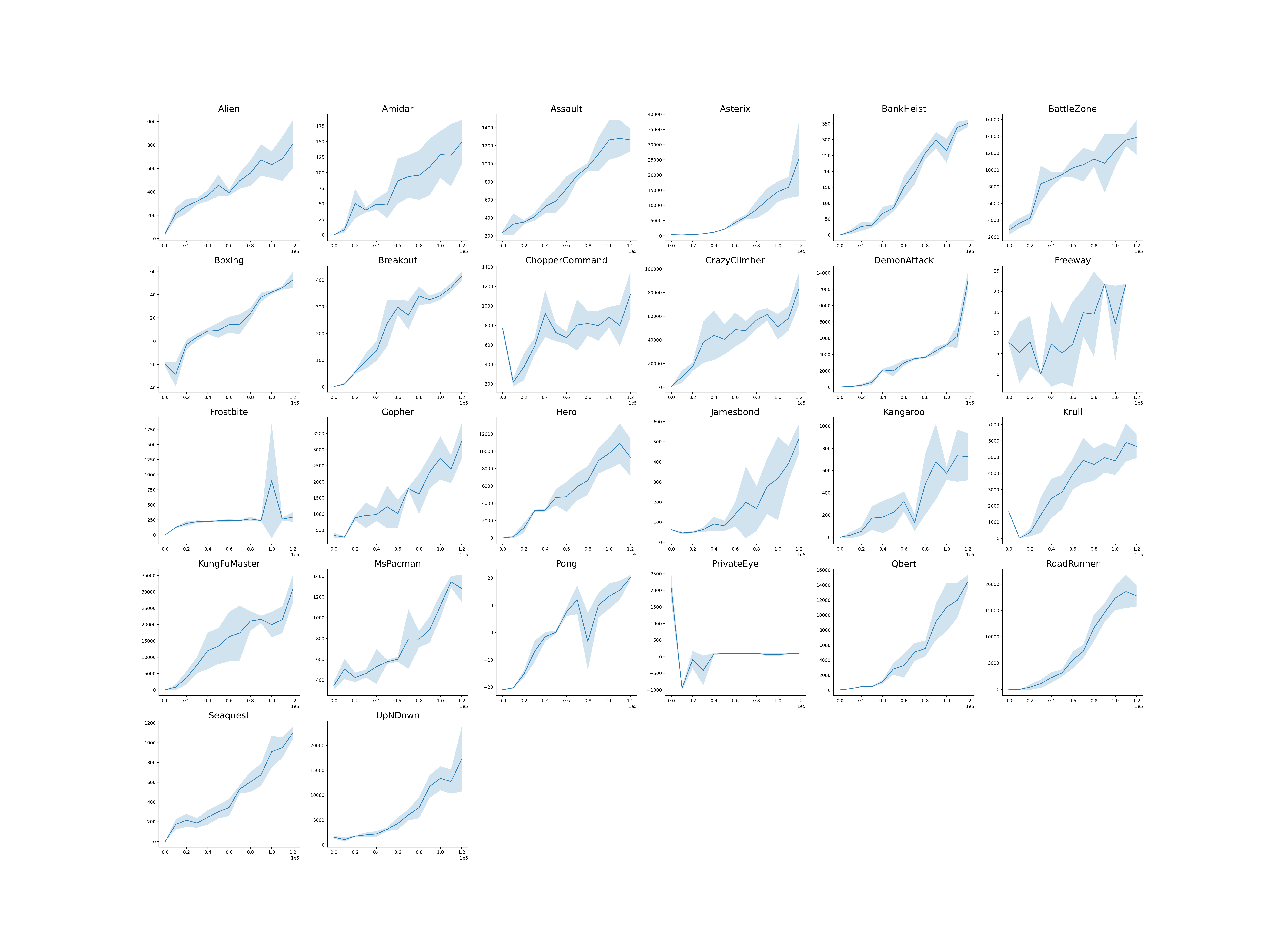}
  \end{center}
  \vskip -0.25cm
  \caption{\textbf{Evaluation curves of \textit{EfficientZero} on Atari 100k benchmark for individual games}. The average of the total rewards among 32 evaluation seeds for 3 runs is show on the y-axis and the number of total training steps is 120,000, shown on the x-axis.}
  \label{fig:final_results}
  \vskip -0.25cm
\end{figure}

We evaluate the EfficientZero on Atari 100k benchmark with a total of 26 games. Here are the evaluation curves during training, as shown in Figure \ref{fig:final_results}.

Besides, we also report the scores for 3 runs (different seeds) with 32 evaluation seeds across the 26 Atari games, which is shown in Table \ref{tab:scores_all_runs}.

\begin{figure}
    \centering
    \includegraphics[width=1\linewidth]{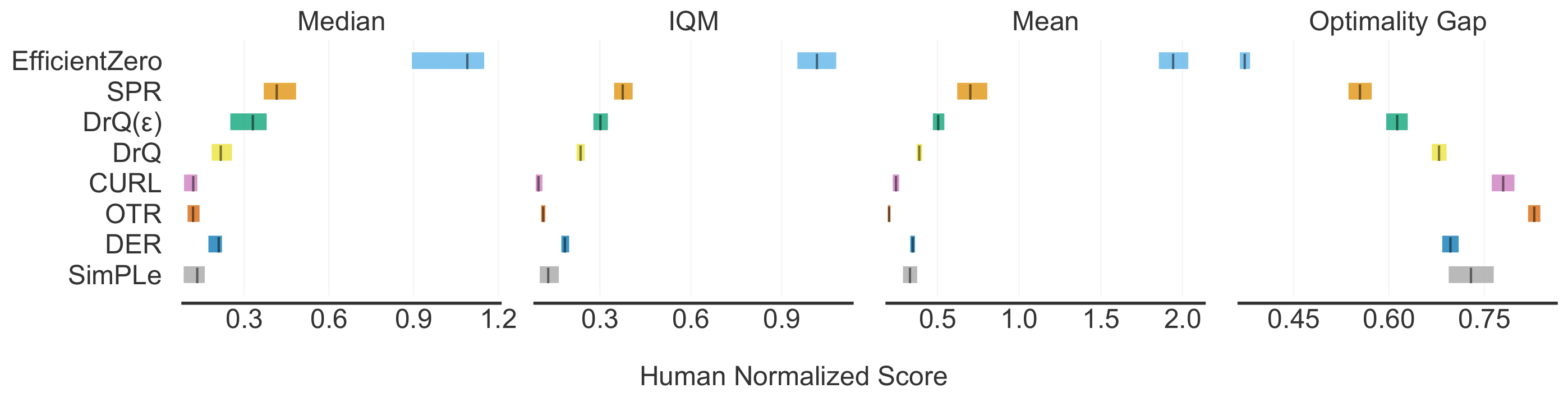}
    \caption{\textbf{Aggregate metrics on Atari 100k benchmark with 95\% CIs}. Here the higher mean, median and IQM scores and lower optimality gap indicate better performance. The CIs are estimated by the percentile bootstrap with stratified sampling. All results except EfficientZero are from \citet{agarwal2021deep}. And all the methods are based on 10 runs per game except SimPLe with 5 runs and EfficientZero with 3 runs. EfficientZero significantly outperforms the other methods concerning the four metrics.}
    \label{fig:aggregate}
\end{figure}

Recently, \citet{agarwal2021deep} propose to use statistical tools to present more robust and efficient aggregate metrics. Here we display the corresponding results based on its open-sourced codebase. Figure \ref{fig:aggregate} illustrates that EfficientZero significantly outperforms the other methods on Atari 100k benchmark concerning all the metrics.

\begin{table}[ht]
    \caption{Scores reported for 3 random seeds for each of the above games, with the last two columns being the mean and standard deviation across the runs. Each run is evaluated with 32 different seeds.}
    \label{tab:scores_all_runs}
\begin{center}
\begin{small}
\centering
\scalebox{0.85}{
\centering
\begin{tabular}{lccc|cc}
\toprule
Game & Seed 0 & Seed 1 & Seed 2 & Mean & Std \\
\midrule
Alien & 1093.1 & 622.2 & 710.3 & 808.5 & 204.4 \\
Amidar & 198.7 & 116.4 & 130.6 & 148.6 & 35.9 \\
Assault & 1436.3 & 1150.8 & 1202.2 & 1263.1 & 124.3 \\
Asterix & 18421.9 & 43220.2 & 15031.3 & 25557.8 & 12565.7 \\
Bank Heist & 362.6 & 336.3 & 354.0 & 351.0 & 10.9 \\
Battle Zone & 11812.5 & 13100.8 & 16700.4 & 13871.2 & 2068.5 \\
Boxing & 45.9 & 49.9 & 62.4 & 52.7 & 7.0 \\
Breakout & 432.8 & 418.7 & 390.9 & 414.1 & 17.4 \\
ChopperCommand & 1190.9 & 1360.9 & 800.0 & 1117.3 & 234.8 \\
Crazy Climber & 98640.2 & 65520.4 & 87660.1 & 83940.2 & 13774.6 \\
Demon Attack & 11517.5 & 14323.3 & 13170.8 & 13003.9 & 1151.5 \\
Freeway & 21.8 & 21.8 & 21.8 & 21.8 & 0.0 \\
Frostbite & 407.1 & 225.5 & 256.3 & 296.3 & 79.4 \\
Gopher & 3002.6 & 2744.2 & 4034.1 & 3260.3 & 557.2 \\
Hero & 12349.1 & 8006.5 & 7592.0 & 9315.9 & 2151.5 \\
Jamesbond & 530.7 & 600.3 & 420.1 & 517.0 & 74.2 \\
Kangaroo & 980.2 & 460.7 & 731.3 & 724.1 & 212.1 \\
Krull & 4839.5 & 5548.5 & 6602.0 & 5663.3 & 724.1 \\
Kung Fu Master & 28493.1 & 36840.7 & 27500.5 & 30944.8 & 4188.7 \\
Ms Pacman & 1465.0 & 1203.4 & 1175.3 & 1281.2 & 130.4 \\
Pong & 20.6 & 18.8 & 21.0 & 20.1 & 1.0 \\
Private Eye & 100.0 & 90.0 & 100.0 & 96.7 & 4.7 \\
Qbert & 15458.1 & 14577.5 & 13310.0 & 14448.5 & 881.7 \\
Road Runner & 17843.8 & 20140.0 & 15270.2 & 17751.3 & 1989.2 \\
Seaquest & 1038.1 & 1078.2 & 1184.4 & 1100.2 & 61.7 \\
Up N Down & 22717.5 & 8095.6 & 20979.4 & 17264.2 & 6521.9 \\
\bottomrule
\end{tabular}
}
\end{small}
\end{center}
\end{table}

\subsection{Open Source EfficientZero Implementation}

\label{sec:opensource}
MCTS-based RL algorithms present a promising future research direction: to achieve strong performance with model-based methods. However, two major practical obstacles prevent them from being widely used currently. First, there are no high-quality open-source implementations of these algorithms. Existing implementations~\cite{muzero-general, muzero-pytorch} can only deal with simple state-based environments, such as CartPole~\cite{barto1983neuronlike}. Accurately scaling to complex image input environments requires non-trivial engineering efforts. Second, MCTS RL algorithms such as MuZero~\cite{schrittwieser2020mastering} require a large number of computations. For example, MuZero needs 64 TPUs to train 12 hours for one agent on Atari games. The high computational costs pose problems both for the future development of such methods as well as practical applications. 

We think our open-source implementation of EfficientZero can drastically accelerate the research in MCTS RL algorithms. Our implementation is computationally friendly. To train an Atari agent for 100k steps, it only needs 4 GPUs to train 7 hours. Our framework could potentially have a large impact on many real-world applications, such as robotics since it requires significantly fewer samples. 

\begin{wrapfigure}{R}{0.65\textwidth}
  \begin{center}
  \vskip -.5cm
    \includegraphics[width=\linewidth]{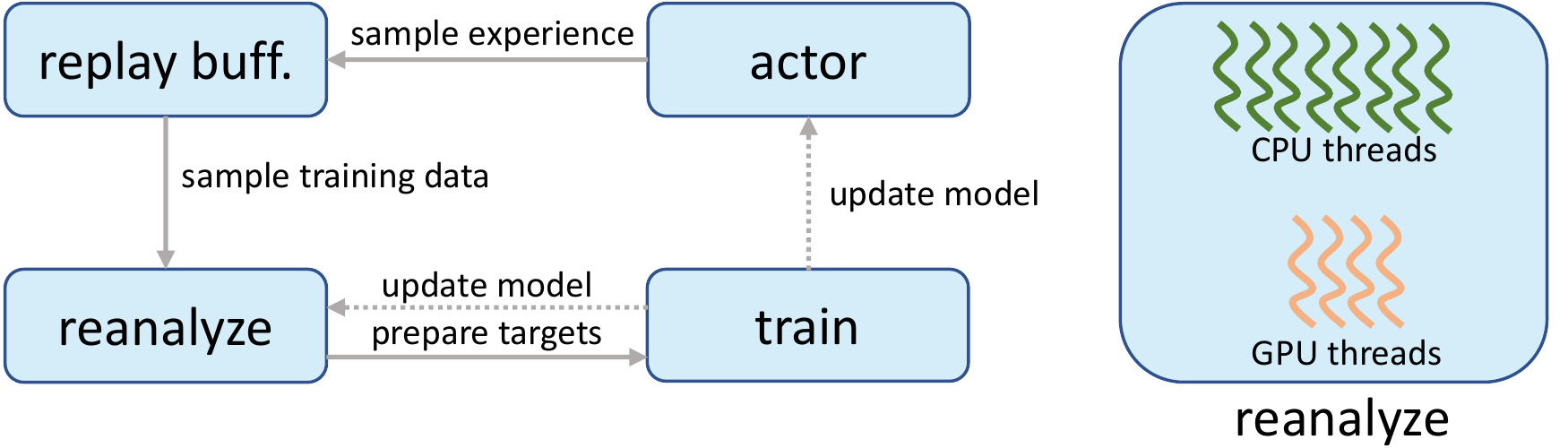}
  \end{center}
  \vskip -0.25cm
  \caption{EfficientZero implementation overview. }
  \label{fig:system}
  \vskip -0.25cm
\end{wrapfigure}

Our open-source framework aims to provide an easy way to understand the implementation while keeping relatively high compute efficiency. As shown in Fig. \ref{fig:system}, the system is composed of four components: the replay buffer, the experience sampling actor, the reanalyze training target preparation module, and the training component. 

To make sure the framework is easy to use, we implement them based on Ray~\cite{moritz2018ray}, and the four components are implemented as ray actors which run in parallel. The main computation bottleneck is in the reanalyze module, which samples from the replay, and runs an MCTS search on each observation. To accelerate the reanalyze module, we split the reanalyze computation into the CPU part and the GPU part, such that computation on CPU and GPU are run in parallel. We use a different number of actors between CPU and GPU to match their total throughput. To increase the throughput on GPU, we also collocate multiple batch computation threads on one GPU, as in \citet{tian2019elf}. We also implement the MCTS in C++ to avoid performance issues with Python on large amounts of atomic computations.

We implement the MCTS by a couple of important techniques, which are quite crucial to improve the efficiency of the MCTS process. On the one hand, we implement batch MCTS to allow the agent to search a batch of trees in parallel, to enlarge the throughput of MCTS during self-play and reanalyzing targets. On the other, we choose C++ in the MCTS process. However, the process of MCTS needs to do searching as well as model inference, which needs to communicate with Pytorch. Therefore, we use Python to do model inference, C++ to do other atomic computations, and Cython to communicate between Python contexts and C++ contexts. In another word, we use pure C++ to do selection, expansion, and backup while using neural networks in Python.
Meanwhile, we build a database to store the hidden states in Python while storing the corresponding data index during the searching process in C++.
For more details of the implementation, please refer to \url{https://github.com/YeWR/EfficientZero}.

\end{document}